\pdfoutput=1
\documentclass[conference]{IEEEtran}
\IEEEoverridecommandlockouts

\usepackage[T1]{fontenc}
\usepackage{cite}
\usepackage{amsmath,amssymb}
\usepackage{pifont}
\renewcommand{\checkmark}{\ding{51}}
\usepackage{graphicx}
\usepackage{booktabs}
\usepackage{multirow}
\usepackage{makecell}
\usepackage{array}
\usepackage[table]{xcolor}
\usepackage[hidelinks]{hyperref}

\definecolor{best}{HTML}{DDEAF9}
\newcommand{\B}[1]{\textbf{#1}}
\newcommand{\grare}{G\textsubscript{Rare}}

\begin{document}

\title{Beyond Multi-Stage Pipelines: A Label-Correlation Ensemble for Multi-Label Retinal Fundus Disease Classification}

\author{
\IEEEauthorblockN{Md Asif Uddin\IEEEauthorrefmark{1},
Jannatul Ferdous,
Asif Islam,
Zobaer Mahmud Zisan,\\
Promit Ghosh,
Shah Ahafik Arman,
Niloy Farhan,
Rafeed Rahman}
\IEEEauthorblockA{Department of Computer Science and Engineering, BRAC University, Dhaka, Bangladesh\\
\{md.asif.uddin, jannatul.ferdous17, asif.islam, zobaer.mahmud.zisan,\\ Promit.ghosh, shah.ahafik.arman\}@g.bracu.ac.bd;\ 
\{niloy.farhan, rahman.rafeed\}@bracu.ac.bd\\
\IEEEauthorrefmark{1}Corresponding author}
}

\maketitle

\begin{abstract}
Retinal fundus images frequently carry more than one disease, and public fundus
datasets are heterogeneous and severely long-tailed. Clinical triage suggests a
staged design, in which a screening gate, a disease-group router and group-specific
classifiers operate in sequence, but it is unclear whether this hierarchy earns its
complexity. We present a controlled comparison of three architectures on a unified
39-target label space built from RFMiD~1.0, RFMiD~2.0 and ODIR-5K (12{,}652 images):
(i) a three-stage gated cascade of eight models, (ii) a flat model with 39
independent sigmoid outputs, and (iii) a label-correlation model that replaces the
flat linear head with 39 learned label queries and a two-layer transformer decoder.
All three share one Swin-V2-B backbone, one preprocessing pipeline and one training
protocol, so differences are attributable to the architecture. Staging offers no
consistent advantage: tracing every missed disease instance to the stage that lost
it shows that the two coarse gates discard 34.2\% of true instances while the
fine-grained classifiers lose only 6.7\%, and the gates' hard zeros depress the
cascade's macro-AUC to 0.747. Label-correlation modeling gives the largest single
gain, raising the detection rate from 0.621 to 0.682 (0.689 with test-time
augmentation). Blending the flat and label-correlation models under four-view
test-time augmentation is the best configuration on six of eight metrics (exact
match 0.694, micro-F1 0.678, macro-F1 0.456, macro-AUC 0.972). On definition-matched class subsets it matches or exceeds four recent studies on
macro- and micro-F1, most clearly on low-support pathologies. We also show that ensembling a gated
cascade with continuous-output models is harmful, and that per-source negative
sampling is needed to remove a dataset shortcut that otherwise yields a spurious
AUC of 1.000.
\end{abstract}


\section{Introduction}
Diabetic retinopathy, glaucoma and age-related macular degeneration are among the
leading causes of preventable blindness; more than 2.2 billion people live with a
vision impairment, nearly half of which could have been prevented or remains
unaddressed~\cite{who2019}. Fundus photography is cheap and non-invasive, and deep
networks reach specialist-level accuracy on single diseases such as diabetic
retinopathy~\cite{gulshan2016,ting2017}. Real screening, however, is a
\emph{multi-label} problem: one eye may show diabetic retinopathy, laser scars and
media haze at once, and the clinically important tail contains dozens of conditions
with only a handful of examples each~\cite{pachade2021,cen2021}.

Two design philosophies compete for this problem. A \emph{hierarchical cascade}
mirrors clinical triage: a binary gate removes healthy eyes, a router assigns the
remaining eyes to broad disease groups, and small group-specific classifiers make
the fine-grained call. Each sub-problem is easier, and failures are attributable to
a stage. A \emph{flat} model predicts every target from one shared representation,
avoiding early irreversible decisions. A third option keeps the flat topology but
lets labels interact, so that evidence for one disease can inform another.

Prior multi-disease fundus studies usually compare across different backbones,
preprocessing and label sets, which makes it impossible to separate the effect of
the architecture from everything else. We hold everything except the architecture
fixed and ask a direct question: \emph{does the hierarchy earn its complexity?}

\textbf{Contributions.}
\begin{itemize}
\item A controlled comparison of a three-stage cascade, a flat classifier and a
label-correlation transformer decoder on a unified 39-target label space merged
from RFMiD~1.0, RFMiD~2.0 and ODIR-5K, with a shared backbone, preprocessing, loss,
optimizer and calibration protocol.
\item A \emph{loss-cascade} analysis that attributes every missed disease instance
to the stage that lost it, showing that the coarse gates, not the fine-grained
classifiers, bound the cascade's recall, and that its hard-gated zeros invalidate
rank-based metrics.
\item A nine-configuration ablation isolating test-time augmentation (TTA),
threshold re-tuning, label-correlation modeling and ensembling, including the
negative result that a gated cascade is a harmful ensemble member.
\item A source-balanced negative-sampling scheme that removes a provenance shortcut
in the media-opacity group, and a subset-matched comparison with four recent
studies.
\item Cleaned code and the full test-set predictions of the best configuration,
publicly released.
\end{itemize}

\section{Related Work}
\textbf{Multi-disease fundus classification.} RFMiD~\cite{pachade2021} and
RFMiD~2.0~\cite{panchal2023} annotate 46 and 49 conditions, and ODIR-5K~\cite{odir}
provides eight categories with per-eye diagnostic keywords. Cen \emph{et
al.}~\cite{cen2021} detected 39 conditions with a hierarchical ensemble trained on a
large private corpus. On public data, Rodr\'iguez \emph{et al.}~\cite{rodriguez2022}
introduced the MuReD dataset and a C-Tran label-query transformer;
Bhati \emph{et al.}~\cite{bhati2022} targeted imbalance with discriminative kernel
convolutions; Singh \emph{et al.}~\cite{singh2025} ensembled CNN and transformer
backbones; and recent work benchmarks twelve CNN, transformer, hybrid and
vision-language models on RFMiD~\cite{dey2026}, or combines dual CNN
backbones~\cite{mahmood2026} and asymmetric losses in a cascade~\cite{saikia2026}.
None isolates the effect of hierarchy against a matched flat model.

\textbf{Hierarchical classification} exploits a label taxonomy by decomposing the
task~\cite{silla2011}; top-down cascades are known to propagate early errors, but
the size of this effect in multi-label medical screening is rarely measured.

\textbf{Label correlation.} ML-GCN~\cite{chen2019} propagates label embeddings over a
co-occurrence graph, while C-Tran~\cite{lanchantin2021} and
Query2Label~\cite{liu2021q2l} use learnable label queries in a transformer decoder.
Our label-correlation model follows the query-decoder design.

\textbf{Imbalance and shortcuts.} Focal~\cite{lin2017} and
asymmetric~\cite{ridnik2021} losses re-weight easy negatives. Models trained on
pooled multi-site data can learn site or device signatures instead of
pathology~\cite{zech2018,geirhos2020}, which motivates our per-source evaluation and
negative sampling. TTA~\cite{shanmugam2021} and deep ensembles~\cite{lakshmi2017}
are standard ways to stabilize predictions.

\section{Data and Label Space}
\begin{table}[t]
\centering
\caption{Merged dataset. Each source keeps its official split.}
\label{tab:data}
\setlength{\tabcolsep}{4pt}
\begin{tabular}{@{}lrrrl@{}}
\toprule
Source & Train & Val & Test & Native labels\\
\midrule
ODIR-5K (per eye)~\cite{odir} & 6{,}092 & 861 & 1{,}642 & 8 categories + keywords\\
RFMiD~1.0~\cite{pachade2021} & 1{,}920 & 640 & 640 & 46 conditions\\
RFMiD~2.0~\cite{panchal2023} & 508 & 177 & 172 & 49 conditions\\
\midrule
Merged & 8{,}520 & 1{,}678 & 2{,}454 & 39 scored targets\\
\quad of which healthy & 3{,}652 & 617 & 1{,}005 & \\
\bottomrule
\end{tabular}
\end{table}

\textbf{Harmonization.} ODIR-5K is converted from patient-level records to per-eye
images whose labels come from that eye's diagnostic keywords. The healthy flags of
the three sources (RFMiD~1.0's inverted \emph{Disease Risk}, RFMiD~2.0's
\emph{WNL}, ODIR's \emph{N}) are unified into one \emph{Normal} indicator. ODIR's
``Other'' category is removed because it is a catch-all rather than a definable
pathology. Image identifiers are namespaced by source to prevent silent collisions,
and the official splits are merged split-wise (Table~\ref{tab:data}); the merged
test split is sealed and read only for final evaluation.

\textbf{Label space.} The 38 named diseases with at least six training images are
organized into six clinically motivated groups (Table~\ref{tab:groups}). The 21
ultra-rare classes with five or fewer training images are pooled into one
\grare{} target that is detected but not sub-classified, giving 39 scored targets.
The distribution is extremely long-tailed: diabetic retinopathy has 2{,}219 training
images, while 16 of the 39 targets have fewer than ten test positives.

\begin{table}[t]
\centering
\caption{Disease groups, training images per group and validation performance of
the Stage-2 router (per group) and the Stage-3 classifiers (macro over the group).}
\label{tab:groups}
\setlength{\tabcolsep}{2.2pt}
\footnotesize
\begin{tabular}{@{}l>{\raggedright\arraybackslash}p{2.35cm}rcccc@{}}
\toprule
& & & \multicolumn{2}{c}{Stage 2} & \multicolumn{2}{c}{Stage 3}\\
\cmidrule(lr){4-5}\cmidrule(l){6-7}
Group & Diseases & Train & AUC & AP & AUC & AP\\
\midrule
G1 Vascular & DR HR HTN BRVO CRVO TV CWS PRH MCA & 2{,}554 & .970 & .969 & .971 & .673\\
G2 Exudative & ARMD DN EDN CSR CME RPEC MHL ERM PT & 698 & .932 & .775 & .926 & .582\\
G3 Optic disc & ODC GLA ODP ODE AION TD ST & 806 & .902 & .745 & .942 & .660\\
G4 Structural & MYA TSLN LS RT RD CF & 694 & .978 & .933 & .983 & .919\\
G5 Media & MH CAT & 656 & .969 & .874 & .983$^\dagger$ & .948$^\dagger$\\
G6 Inflamm. & CRS RS MS AH RP & 162 & .946 & .587 & .951 & .858\\
\grare{} & 21 pooled classes & 42 & .878 & .177 & -- & --\\
\bottomrule
\multicolumn{7}{@{}p{\columnwidth}@{}}{\footnotesize $^\dagger$With source-balanced negatives; the positives-only model scores 1.000 / 1.000 through a source shortcut.}
\end{tabular}
\end{table}

\section{Methods}
\begin{figure*}[t]
\centering
\includegraphics[width=\textwidth]{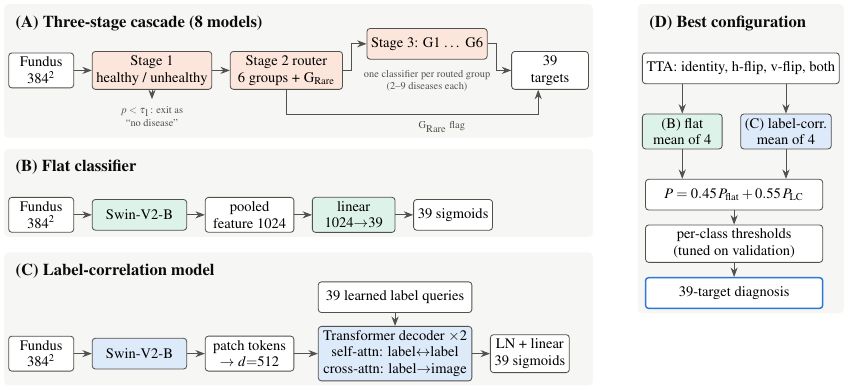}
\caption{The four systems compared. (A) Healthy images exit at Stage~1; unhealthy
images are routed to every group that fires, and each routed group's classifier
makes the fine-grained call. (B) One linear head predicts all 39 targets. (C) 39
label queries attend to the image tokens and to each other. (D) The best
configuration blends (B) and (C) under four-view TTA.}
\label{fig:arch}
\end{figure*}

\subsection{Problem formulation}
Each image $x$ has a target vector $\mathbf{y}\in\{0,1\}^{39}$; a healthy image has
$\mathbf{y}=\mathbf{0}$, and an image may carry several diseases. Every system outputs
scores $p_c(x)\in[0,1]$ and decisions $\hat y_c=\mathbf{1}[p_c(x)\ge\tau_c]$. For the
cascade, let $s_1(x)$ be the Stage-1 probability of disease, $s_{2,g}(x)$ the router
score of group $g$ and $s_{3,c}(x)$ the Stage-3 score of disease $c\in g$. With hard
gating,
\begin{equation}
p^{\mathrm{casc}}_c(x)=\mathbf{1}\!\left[s_1(x)\ge\tau_1\right]\,
\mathbf{1}\!\left[s_{2,g(c)}(x)\ge\tau_{g(c)}\right]\, s_{3,c}(x),
\label{eq:cascade}
\end{equation}
so any gate error forces $p_c=0$ irrespective of the Stage-3 evidence. The flat and
label-correlation models instead produce $p_c(x)=\sigma(f_c(x))$ for every $c$ and
every image. This difference drives both the recall ceiling of the cascade
(Section~\ref{sec:loss}) and the behaviour of its rank-based metrics.

\subsection{Shared protocol}
All models use Swin-V2-B~\cite{liu2022swinv2} with ImageNet-22k pre-training and
ImageNet-1k fine-tuning, loaded through timm~\cite{timm},\footnote{Checkpoint
\texttt{swinv2\_base\_window12to24\_\allowbreak 192to384.\allowbreak ms\_in22k\_ft\_in1k}.} at $384\times384$. Preprocessing crops
each image to the bounding box of the illuminated retina and zero-pads the short side
to a square, so no retinal tissue is discarded. Training augmentation is random
resized crop (scale 0.85--1.0), horizontal and vertical flips, $\pm15^\circ$ rotation
and mild color jitter. Multi-label models use binary cross-entropy with a
per-class positive weight $w_c=\min(\max(N^-_c/N^+_c,1),50)$. Optimization is
AdamW~\cite{loshchilov2019} (weight decay $10^{-2}$) with linear warm-up and cosine
decay, mixed precision, gradient clipping at 1.0, batch size 4 with gradient
accumulation, and early stopping on the validation monitor. For the 39-target models, decision thresholds are
tuned per class on \emph{validation} by maximizing F1 over $\{0.05,0.075,\dots,0.95\}$
for every target with at least ten validation positives (22 of 39); the others use
0.5. The test split is used once. Table~\ref{tab:hparams} lists the per-model
settings.

\begin{table}[t]
\centering
\caption{Training settings. All models: Swin-V2-B at 384\,px, AdamW, weight decay
$10^{-2}$, warm-up then cosine decay, batch size 4 with gradient accumulation
(Batch: effective batch). $^s$Softmax; all other heads are sigmoid.}
\label{tab:hparams}
\setlength{\tabcolsep}{2.4pt}
\footnotesize
\begin{tabular}{@{}lccccccc@{}}
\toprule
Model & Out. & Images & LR & Batch & Ep. & Pat. & Monitor\\
\midrule
Stage 1 & 2$^s$ & 8{,}520 & $10^{-5}$ & 8 & 60 & 8 & AUC\\
Stage 2 & 7 & 4{,}868 & $2{\cdot}10^{-5}$ & 8 & 30 & 7 & macro-AUC\\
Stage 3 ($\times$6) & 2--9 & 162--2{,}554 & $2{\cdot}10^{-5}$ & 8 & 30 & 7 & macro-AP\\
Flat & 39 & 8{,}520 & $2{\cdot}10^{-5}$ & 8 & 30 & 7 & macro-AP\\
Label-corr. & 39 & 8{,}520 & $2{\cdot}10^{-5}$ & 16 & 25 & 6 & macro-AP\\
\bottomrule
\end{tabular}
\end{table}

\subsection{Architecture A: three-stage cascade}
\textbf{Stage~1} is a two-way softmax screening gate (healthy vs.\ unhealthy)
trained on all 8{,}520 training images with class-weighted cross-entropy (learning
rate $10^{-5}$, 60 epochs, patience 8; class weighting outperformed a weighted
sampler, validation AUC 0.938 vs.\ 0.920). \textbf{Stage~2} is a multi-label router
with seven sigmoid outputs (six groups and \grare{}) trained on the 4{,}868 unhealthy
training images; an image is routed to every group whose score exceeds its tuned
threshold. Among BCE, focal, asymmetric loss and GeM pooling, plain BCE gave the best
validation macro-AP (0.722 vs.\ 0.703, 0.714, 0.710). \textbf{Stage~3} comprises six
group-specific multi-label classifiers trained with ground-truth routing on the
images of their group (2--9 classes each). \grare{} is decided at Stage~2 because
its constituent classes are too scarce for a dedicated model. Gating is hard: an
image rejected at Stage~1, or not routed to a group, receives probability zero for
every downstream target, exactly as a deployed cascade behaves.

\textbf{Source-balanced negatives.} In G5, media haze comes almost entirely from
RFMiD and cataract almost entirely from ODIR, so a positives-only G5 model reaches a
validation AUC of 1.000 by recognizing the camera. We therefore add negatives
(unhealthy images without a G5 disease) sampled \emph{per source} in proportion to
that source's positives, which equalizes the positive rate within every source. The
de-confounded model scores a macro-AUC of 0.983; measured within single sources,
where provenance cannot explain the result, the AUC is 0.987 for cataract in ODIR
and 0.954 for media haze in RFMiD~1.0. This model is used in the cascade.

\subsection{Architecture B: flat classifier}
A single Swin-V2-B maps its 1{,}024-d pooled feature through a linear layer to 39
independent sigmoids. It is trained on all training images (healthy images have
all-zero targets) and checkpointed on validation macro-AP.

\subsection{Architecture C: label-correlation model}
The flat linear head is replaced by a query decoder. Swin patch tokens are
projected to $d=512$ and serve as decoder memory. Thirty-nine learned query vectors,
one per target, pass through a two-layer pre-norm transformer decoder (8 heads,
feed-forward width 2{,}048, dropout 0.1); self-attention lets the labels exchange
evidence and cross-attention lets each query inspect the image regions relevant to
its disease. A LayerNorm and a shared linear layer map each query to one logit
(95.8\,M parameters vs.\ 86.9\,M for the flat model). The training set contains 204
distinct co-occurring disease pairs (e.g., DR--HR, TSLN--MH, ODC--TSLN, DR--LS),
which an independent sigmoid head cannot represent. Training is identical to the
flat model except for a larger effective batch (16) and a shorter schedule (Table~\ref{tab:hparams}). Both early-stopped after 19 epochs; the selected checkpoints have validation macro-AP 0.516 (flat) and 0.512 (label-correlation).

\subsection{Architecture D: TTA and ensembling}
TTA averages sigmoid outputs over four label-preserving views: identity,
horizontal flip, vertical flip and both. Because averaging shifts the probability
distribution, thresholds are re-tuned on validation TTA outputs. The ensemble is
$P = w\,P_{\mathrm{LC}} + (1-w)\,P_{\mathrm{flat}}$, with $w$ swept over
$\{0,0.05,\dots,1\}$ to maximize validation macro-AP; the selected $w=0.55$ is
applied unchanged to test, followed by per-class thresholds tuned on the blended
validation outputs.

\subsection{Metrics}
We report exact-match (subset) accuracy, mean per-image Jaccard index, micro- and
macro-F1, macro-F1 over the 23 well-supported targets ($\geq10$ test positives),
macro-AUC, macro-AP and the \emph{detection rate}, the fraction of all 1{,}719 true
disease instances in the test split that are predicted positive. Macro averages run
over the targets with at least one test positive. Every configuration is scored by
the same routine on the same split.

\section{Results}
\subsection{Where the cascade loses disease}\label{sec:loss}
\begin{figure}[t]
\centering
\includegraphics[width=\columnwidth]{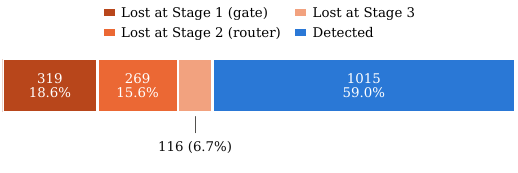}
\caption{Loss cascade: the stage at which each true disease instance of the test
split is lost. The two coarse gates account for 34.2\% of all instances.}
\label{fig:cascade}
\end{figure}

Table~\ref{tab:cascade} and Fig.~\ref{fig:cascade} trace the cascade on the test
split. Stage~1 reaches 0.852 accuracy and 0.928 AUC but sends 297 diseased images
home as healthy, and every disease they carry is lost. Stage~2 is accurate on the
images it receives (macro-AUC 0.955), yet a missed group makes all of its diseases
unreachable. Of the 1{,}719 true disease instances, 34.2\% are lost at the two coarse
gates (18.6\% at Stage~1, 15.6\% at Stage~2) and only 6.7\% by the fine-grained
classifiers; 59.0\% are detected. The coarse decisions, not the fine-grained
ones, bound the cascade's recall. The Stage-1 threshold trades this loss against healthy recall
(Table~\ref{tab:s1}): moving it from 0.5 to 0.2 raises unhealthy recall from 0.845
to 0.876 at the cost of healthy recall (0.912 to 0.853), but even the most liberal
setting misses about one diseased image in ten, and every miss removes all of that
image's diseases.
\begin{table}[t]
\centering
\caption{Stage-1 operating point on validation (1{,}061 unhealthy, 617 healthy
images).}
\label{tab:s1}
\setlength{\tabcolsep}{6pt}
\footnotesize
\begin{tabular}{@{}lcccccc@{}}
\toprule
Threshold $\tau_1$ & .10 & .20 & .30 & .50 & .70 & .90\\
\midrule
Unhealthy recall & .896 & .876 & .862 & .845 & .825 & .770\\
Healthy recall & .799 & .853 & .887 & .912 & .940 & .968\\
Accuracy & .861 & .867 & .871 & .870 & .867 & .843\\
\bottomrule
\end{tabular}
\end{table}

\begin{table}[t]
\centering
\caption{Three-stage cascade on the test split ($n=2{,}454$), all upstream errors
propagated.}
\label{tab:cascade}
\setlength{\tabcolsep}{4pt}
\begin{tabular}{@{}lr@{}}
\toprule
Stage~1 accuracy / AUC & .852 / .928\\
Stage~1 sensitivity / specificity & .795 / .933\\
Diseased images rejected at Stage~1 & 297 of 1{,}449\\
Stage~2 macro-AUC on received images & .955\\
Exact group set / exact disease set & .732 / .687\\
\midrule
Images: correct healthy / exact set / partial & 942 / 744 / 213\\
Images: total miss / false alarm & 492 / 63\\
\midrule
Instances lost at Stage 1 / 2 / 3 & 319 / 269 / 116\\
Instances detected & 1{,}015 / 1{,}719 (59.0\%)\\
\bottomrule
\end{tabular}
\end{table}

\textbf{Gated zeros invalidate rank metrics.} The cascade's macro-AUC is 0.747
although its six clinical-group outputs and its six group classifiers all exceed
0.90 macro-AUC on validation (Table~\ref{tab:groups}).
A true positive discarded at a gate receives a hard score of 0.000 and is ranked
below every true negative with a small non-zero score, which breaks the probability
ordering that AUC and AP measure. Decision-based metrics (F1, exact match,
detection) remain meaningful and are the appropriate basis for comparing a gated
system.

\textbf{Case studies.} Table~\ref{tab:cases} traces individual test images through
both designs. In \texttt{4719\_right}, Stage~1 assigns $s_1=0.243$ to an eye with
diabetic retinopathy and the image exits as healthy; the flat model also misses it,
so this is a hard image rather than a pure gating error. In
\texttt{rfmid\_1\_20\_test}, the router fires on the vascular and structural groups
but not on the exudative group, so the macular hole becomes unreachable, the
structural classifier finds the laser scars but not the tessellation, and the
vascular classifier adds a false diabetic retinopathy; the flat model finds the
laser scars without the false positive. Both designs flag glaucoma on the healthy
eye \texttt{3903\_left}, and both are exactly right on the multi-label image
\texttt{rfmid\_1\_240\_test}.

\begin{table}[t]
\centering
\caption{Case studies on test images (\checkmark{} found, $\times$ missed,
FP false positive).}
\label{tab:cases}
\setlength{\tabcolsep}{3pt}
\footnotesize
\begin{tabular}{@{}>{\raggedright\arraybackslash}p{1.45cm}>{\raggedright\arraybackslash}p{1.2cm}>{\raggedright\arraybackslash}p{3.15cm}>{\raggedright\arraybackslash}p{1.55cm}@{}}
\toprule
Image & Truth & Cascade & Flat\\
\midrule
4719\_right & DR & exits at Stage 1 ($s_1{=}0.24$): DR $\times$ & DR $\times$\\
rfmid\_1\_20 & MHL, TSLN, LS & G2 not routed: MHL $\times$; TSLN $\times$ at Stage 3; LS \checkmark; DR FP & LS \checkmark; MHL $\times$; TSLN $\times$\\
3903\_left & healthy & $s_1{=}0.61$, routed to G3: GLA FP & GLA FP\\
rfmid\_1\_240 & TSLN, MH & TSLN \checkmark, MH \checkmark & TSLN \checkmark, MH \checkmark\\
2322\_right & CAT & CAT \checkmark & CAT \checkmark\\
\bottomrule
\end{tabular}
\end{table}

\begin{table*}[t]
\centering
\caption{Ablation on the test split (2{,}454 images, 39 targets). All thresholds
and blend weights are selected on validation. F1\textsubscript{well}: macro-F1 over
the 23 targets with $\geq 10$ test positives. Best value per column in bold.}
\label{tab:ablation}
\setlength{\tabcolsep}{5.2pt}
\begin{tabular}{llcccccccc}
\toprule
& Configuration & Exact & Jaccard & micro-F1 & macro-F1 & F1\textsubscript{well} & macro-AUC & macro-AP & Detection\\
\midrule
1 & Flat baseline & .6789 & .7229 & .6642 & .4426 & .5767 & .9430 & .5048 & .6207\\
2 & Flat + TTA$\times$4 & .6817 & .7248 & .6637 & .4420 & .5707 & .9470 & .5144 & .6056\\
3 & Flat + TTA$\times$4 + re-tuned thresholds & .6764 & .7273 & .6653 & .4444 & .5748 & .9470 & .5144 & .6492\\
4 & Three-stage cascade & .6870 & .7278 & .6495 & .3917 & .5248 & .7467$^*$ & .3713$^*$ & .5939\\
5 & Cascade + flat ensemble ($w=0.5$) & .6817 & .7248 & .6543 & .4154 & .5615 & .9500 & .5213 & .5934\\
6 & Label-correlation & .6679 & .7239 & .6577 & .4518 & .5573 & .9662 & .4599 & .6818\\
7 & Label-correlation + TTA$\times$4 & .6756 & .7302 & .6560 & .4417 & .5568 & .9674 & .4834 & \B{.6894}\\
8 & Three-way ensemble (equal weights) & .6838 & .7229 & .6455 & .4164 & .5220 & .9710 & \B{.5307} & .5736\\
\rowcolor{best}
9 & Flat + label-corr. ensemble + TTA ($w_{\mathrm{LC}}=0.55$) & \B{.6944} & \B{.7399} & \B{.6784} & \B{.4559} & \B{.5942} & \B{.9716} & .5222 & .6376\\
\bottomrule
\multicolumn{10}{@{}p{0.97\textwidth}@{}}{\footnotesize $^*$Depressed by hard-gated zero probabilities. Row 4 scores the cascade's
probabilities at 0.5; its native per-class decisions give micro-F1 .6451, macro-F1 .3922 and detection .5905.}
\end{tabular}
\end{table*}

\subsection{Does the hierarchy earn its complexity?}
Table~\ref{tab:ablation} compares all nine configurations. Against the flat
baseline (row 1), the cascade (row 4) wins slightly on the two image-level metrics,
exact match (0.687 vs.\ 0.679) and Jaccard (0.728 vs.\ 0.723), because its screening
gate cleanly passes healthy eyes. It loses on every instance-level metric: micro-F1
0.649 vs.\ 0.664, macro-F1 0.392 vs.\ 0.443 and detection 0.594 vs.\ 0.621, with
the gap widest on the macro metrics that weight rare classes equally. With eight
models to train, calibrate and maintain, the cascade buys interpretability of its
failure modes, not accuracy.

\begin{figure}[t]
\centering
\includegraphics[width=\columnwidth]{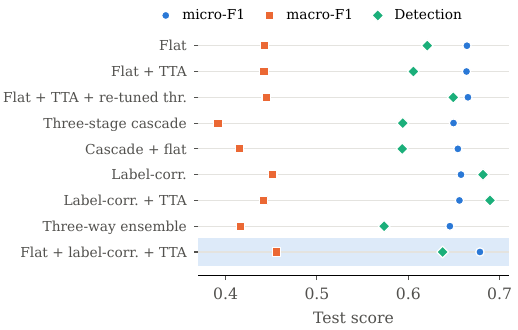}
\caption{Ablation at a glance (values of Table~\ref{tab:ablation}). The shaded row
is the best configuration; the label-correlation rows carry the highest detection.}
\label{fig:ablation}
\end{figure}

\subsection{What each technique contributes}
Fig.~\ref{fig:ablation} summarizes the ablation. \textbf{TTA} applied naively slightly reduces detection ($-0.015$, row 2) because
the averaged probabilities no longer match thresholds tuned on single-view outputs;
with thresholds re-tuned on validation (row 3) it improves detection by $+0.029$ and
micro-F1 by $+0.001$. Re-tuning is a necessary part of TTA, not an option.

\textbf{Label-correlation modeling} is the largest single gain. With nothing changed
but the head, detection rises from 0.621 to 0.682 (row 6) and to 0.689 with TTA
(row 7, the highest of the study), and macro-AUC rises from 0.943 to 0.966.
Detection had been the weakest metric of every flat configuration; co-occurrence
modeling addresses it directly, consistent with its mechanism (evidence of
diabetic retinopathy makes exudates and cotton-wool spots more findable).

\textbf{Ensembling} is additive only between continuous-output models. Blending the
precision-oriented flat model with the recall-oriented label-correlation model
(row 9) is best on six of eight metrics: exact match 0.694, Jaccard 0.740, micro-F1
0.678, macro-F1 0.456, well-supported macro-F1 0.594 and macro-AUC 0.972. In
contrast, averaging the cascade with the flat model (row 5) is worse than its flat partner
(row 3) on micro-F1 (0.654 vs.\ 0.665) and detection (0.593 vs.\ 0.649), and the
three-way ensemble (row 8) is worse still (micro-F1 0.645, detection 0.574). The
cascade's sparse output matrix suppresses its partners' probabilities precisely on
the diseased images its gates discarded. We report this as a negative result:
gated pipelines are unsuitable members of probability-averaging ensembles.

\begin{figure*}[t]
\centering
\includegraphics[width=\textwidth]{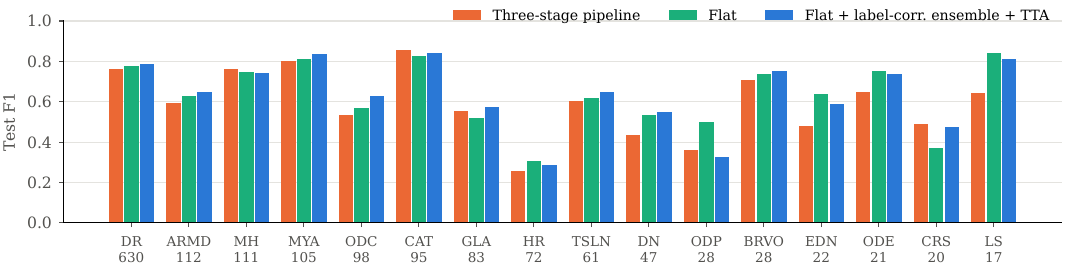}
\caption{Test F1 for the 16 most frequent targets (number of test positives under
each code).}
\label{fig:perdisease}
\end{figure*}

\begin{table*}[t]
\centering
\caption{Per-target test results, ordered by number of test positives ($n$). F1 of the
three-stage cascade (native decisions), the flat model and the best configuration
(flat + label-correlation ensemble + TTA); AP and decision threshold of the best
configuration (``--'': default 0.5, fewer than ten validation positives). Best F1 per
target in bold.}
\label{tab:pertarget}
\setlength{\tabcolsep}{4.2pt}
\footnotesize
\begin{tabular}{@{}lrccccc|lrccccc@{}}
\toprule
Target & $n$ & Cascade & Flat & Best & AP & Thr. & Target & $n$ & Cascade & Flat & Best & AP & Thr.\\
\midrule
DR & 630 & .762 & .778 & \B{.785} & .880 & .70 & RT & 13 & .316 & .375 & \B{.421} & .367 & .35\\
ARMD & 112 & .594 & .627 & \B{.650} & .657 & .40 & CRVO & 11 & .632 & \B{.667} & .667 & .669 & .72\\
MH & 111 & \B{.764} & .745 & .745 & .854 & .90 & CME & 10 & .286 & .444 & \B{.588} & .636 & --\\
MYA & 105 & .802 & .812 & \B{.838} & .902 & .68 & TV & 8 & .000 & .182 & \B{.235} & .276 & --\\
ODC & 98 & .532 & .571 & \B{.627} & .687 & .80 & CWS & 8 & .154 & .400 & \B{.500} & .363 & --\\
CAT & 95 & \B{.856} & .829 & .840 & .922 & .55 & TD & 8 & \B{.375} & .182 & .154 & .211 & --\\
GLA & 83 & .555 & .522 & \B{.573} & .584 & .95 & PRH & 7 & \B{.250} & .000 & .000 & .300 & --\\
HR & 72 & .255 & \B{.309} & .286 & .286 & .37 & AH & 6 & .286 & .444 & \B{.571} & .604 & --\\
TSLN & 61 & .602 & .619 & \B{.646} & .684 & .85 & ERM & 6 & .000 & .222 & \B{.400} & .353 & --\\
DN & 47 & .436 & .532 & \B{.549} & .555 & .85 & AION & 5 & \B{.222} & .000 & .000 & .067 & --\\
ODP & 28 & .361 & \B{.500} & .327 & .386 & .55 & RPEC & 5 & .000 & .000 & \B{.167} & .155 & --\\
BRVO & 28 & .706 & .737 & \B{.750} & .797 & .88 & MHL & 4 & .190 & \B{.400} & .286 & .291 & --\\
EDN & 22 & .481 & \B{.640} & .588 & .700 & .12 & PT & 4 & \B{.400} & .000 & .000 & .126 & --\\
ODE & 21 & .647 & \B{.750} & .737 & .760 & .62 & RP & 3 & .000 & \B{.500} & \B{.500} & .867 & --\\
CRS & 20 & \B{.489} & .370 & .474 & .540 & .57 & ST & 3 & .000 & .000 & .000 & .016 & --\\
LS & 17 & .643 & \B{.839} & .812 & .813 & .57 & HTN & 2 & .000 & \B{.667} & .500 & .583 & --\\
CSR & 16 & .444 & \B{.562} & .552 & .632 & .65 & MCA & 2 & \B{.667} & .000 & .000 & .550 & --\\
\grare{} & 15 & .261 & \B{.364} & .364 & .336 & .72 & CF & 2 & .000 & .000 & .000 & .044 & --\\
RS & 15 & .667 & .560 & \B{.692} & .743 & .72 & RD & 2 & .571 & \B{1.00} & .800 & 1.00 & --\\
MS & 14 & .091 & .111 & \B{.154} & .173 & .55 & & & & & & &\\
\bottomrule
\end{tabular}
\end{table*}

\subsection{Per-disease and per-source behaviour}
Fig.~\ref{fig:perdisease} and Table~\ref{tab:pertarget} show that the best configuration leads or ties on most
frequent targets; the cascade remains competitive on cataract (CAT), media haze (MH)
and chorioretinitis (CRS), where its group specialists see a narrower problem.
The best configuration exceeds an F1 of 0.80 on cataract (0.840), myopia (0.838) and
laser scars (0.812) and reaches 0.785 on diabetic retinopathy, but hypertensive
retinopathy (0.286) and macular scar (0.154) remain hard, and six ultra-low-support
targets (PRH, AION, PT, ST, MCA, CF; $\leq 7$ test positives) are never correctly detected.
Performance also depends on the source (Table~\ref{tab:source}): both models
score best on ODIR-5K, whose eight-category labels are coarse, and worst on
RFMiD~2.0, whose 172 test images carry the richest and rarest label sets.

\begin{table}[t]
\centering
\caption{Test performance by source dataset.}
\label{tab:source}
\setlength{\tabcolsep}{3.2pt}
\footnotesize
\begin{tabular}{@{}lrcccccc@{}}
\toprule
& & \multicolumn{2}{c}{Flat} & \multicolumn{3}{c}{Three-stage cascade}\\
\cmidrule(lr){3-4}\cmidrule(l){5-7}
Source & $n$ & macro-F1 & Exact & macro-F1 & Exact & Detection\\
\midrule
ODIR-5K & 1{,}642 & .625 & .746 & .614 & .757 & .611\\
RFMiD 1.0 & 640 & .375 & .566 & .335 & .567 & .602\\
RFMiD 2.0 & 172 & .319 & .459 & .272 & .465 & .447\\
\bottomrule
\end{tabular}
\end{table}

\textbf{Two operating points.} The label-correlation model with TTA is the
high-sensitivity choice for population screening, where a missed disease costs
more than a false alarm (detection 0.689). The flat + label-correlation ensemble is
the balanced choice for decision support, with the best exact match and F1.

\subsection{Comparison with recent studies}
Our taxonomy matches no published class set, so we compare on
\emph{definition-matched} subsets: for each study we take the identically defined
classes, recompute our metrics on exactly that subset from the released test-set
predictions, and compare with the published figures (Table~\ref{tab:lit}). These
comparisons are computed once on the sealed test split and are not used for model
selection.

\begin{table}[t]
\centering
\caption{Subset-matched comparison (ours / theirs). $n$: shared classes.}
\label{tab:lit}
\setlength{\tabcolsep}{2.5pt}
\footnotesize
\begin{tabular}{@{}llcccc@{}}
\toprule
Study & Data & $n$ & macro-F1 & micro-F1 & macro-AP\\
\midrule
C-Tran~\cite{rodriguez2022} & MuReD & 15 & \B{.657} / .638 & -- & \B{.704} / .685$^a$\\
Dual CNN~\cite{mahmood2026} & MuReD & 15 & \B{.657} / .403 & \B{.726} / .548 & \B{.704} / .582$^b$\\
ASL cascade~\cite{saikia2026} & RFMiD & 26 & \B{.478} / .476 & \B{.702} / .626 & .530 / \B{.539}\\
12 models~\cite{dey2026} & RFMiD & 26 & \B{.478} / .340 & \B{.702} / .509 & --\\
\bottomrule
\multicolumn{6}{@{}p{\columnwidth}@{}}{\footnotesize $^a$Averaged over the comparator's full label set. $^b$Micro PR-AUC. For~\cite{dey2026}, the best of twelve models per metric.}
\end{tabular}
\end{table}

The best configuration matches or exceeds every comparator on macro-F1 and micro-F1
wherever they are reported; the margins are largest against the dual-CNN model~\cite{mahmood2026} (+0.254
macro-F1, +0.178 micro-F1) and the twelve-model benchmark~\cite{dey2026} (+0.138
macro-F1, +0.193 micro-F1), and smallest against the asymmetric-loss
cascade~\cite{saikia2026}, with which macro-F1 and macro-AP are
essentially tied. 
\begin{table}[t]
\centering
\caption{Per-class F1 against C-Tran~\cite{rodriguez2022} on the 15 shared classes
($n$: our test positives).}
\label{tab:ctran}
\setlength{\tabcolsep}{6pt}
\footnotesize
\begin{tabular}{@{}lrccc@{}}
\toprule
Class & $n$ & Ours & C-Tran & $\Delta$F1\\
\midrule
DR & 630 & .785 & .859 & $-$0.074\\
MH & 111 & .745 & .724 & $+$0.021\\
ODC & 98 & .627 & .703 & $-$0.076\\
TSLN & 61 & .646 & .787 & $-$0.141\\
ARMD & 112 & .650 & .615 & $+$0.035\\
DN & 47 & .549 & .607 & $-$0.058\\
MYA & 105 & .838 & .872 & $-$0.034\\
BRVO & 28 & .750 & .867 & $-$0.117\\
ODP & 28 & .327 & .000 & $+$0.327\\
CRVO & 11 & .667 & .571 & $+$0.096\\
RS & 15 & .692 & .706 & $-$0.014\\
ODE & 21 & .737 & .870 & $-$0.133\\
LS & 17 & .812 & .526 & $+$0.286\\
CSR & 16 & .552 & .500 & $+$0.052\\
CRS & 20 & .474 & .364 & $+$0.110\\
\bottomrule
\end{tabular}
\end{table}
The class-level breakdown available for C-Tran (Table~\ref{tab:ctran}) shows where the
advantage comes from: we lead on low-support classes such as optic disc pallor
(ODP, +0.327 F1, where C-Tran detects none), laser scars (+0.286), chorioretinitis
(+0.110) and CRVO (+0.096), while C-Tran leads on frequent, strongly co-occurring
classes such as tessellation ($-0.141$), optic disc edema ($-0.133$) and BRVO
($-0.117$). Because the comparators use different test sets (MuReD's merged
ARIA/STARE/RFMiD partition, or the 640-image RFMiD test set) and one is a
non-peer-reviewed preprint~\cite{saikia2026}, these results situate our model
rather than rank it. Our macro-AUC (0.972 over all 39 targets) cannot be restricted
to the matched subsets and is not compared.

\section{Discussion and Limitations}
\textbf{Staged versus end-to-end design.} The cascade and the flat model are
numerically close overall but behave very differently. The cascade decouples the
39-target problem into local decisions: each Stage-3 classifier searches a space of
2--9 classes, the router follows a clinical taxonomy, and every error can be
localized to a gate, which is valuable for auditing. Its stages can be retrained or
replaced independently. The price is eight networks to train, calibrate and
maintain, and above all the sequential gating loss of Fig.~\ref{fig:cascade}: an
instance discarded upstream is unrecoverable. End-to-end models score every image
against every target with one or two networks, which removes gating losses and
simplifies calibration and deployment, at the cost of intermediate interpretability
and of retraining the whole model for any change.

\textbf{Why the ensemble works.} The two members have opposite biases. The
label-correlation model is recall-oriented: it fires on learned co-occurrence and
reaches the highest detection of the study, occasionally producing false positives
when a secondary condition resembles a primary one. The flat model evaluates each
target independently and is more conservative. With weight 0.55 on the
label-correlation model, the blend keeps most of its recall gain on rare and
co-occurring diseases, while the flat component suppresses spurious activations,
which explains why exact match and every F1 measure improve together.

\textbf{Shortcuts in pooled data.} Merging datasets enlarges the training set but
couples labels to provenance. The perfect G5 score of a positives-only model is a
pure artifact, and it is invisible in pooled validation metrics. Reporting
within-source metrics and sampling negatives per source should be standard practice
whenever a label is concentrated in one source.

\textbf{Limitations.} The ultra-rare classes remain unsolved across all architectures
and losses, which points to data scarcity, not model capacity, as the binding
constraint. Sixteen of the 39 targets have fewer than ten test
positives, so their per-class figures are statistically unstable. The merged test
split is dominated by ODIR-5K, on which all models score best, so aggregate figures
are not directly comparable with single-dataset studies. The intermediate disease
groups follow the lesion literature but were not reviewed by an ophthalmologist.
The cascade's components received more ablation (losses, pooling, samplers) than
the flat and label-correlation models, which were each trained once, an asymmetry
that favors the cascade. Ensemble weights and thresholds are tuned on the same
validation split that selects checkpoints, so a small optimistic bias cannot be
excluded. Future work should replace hard gates with soft probabilistic routing,
bring label-correlation decoding into the router, and address the rare tail with
multi-center data or generative augmentation.

\section{Conclusion}
Under a controlled protocol on 39 retinal targets, a three-stage cascade offers no
consistent advantage over a flat classifier: its coarse gates lose 34.2\% of true
disease instances, five times more than its fine-grained classifiers, and its
hard zeros make rank-based metrics invalid. Replacing the flat head with a
label-query transformer decoder gives the largest single improvement, and blending
it with the flat model under four-view TTA yields the best overall system (micro-F1
0.678, macro-F1 0.456, macro-AUC 0.972). Gated pipelines should not be ensembled
with continuous-output models, and pooled multi-source data must be checked
within each source for provenance shortcuts.

\section*{Data and Code Availability}
All datasets are public (RFMiD~1.0, RFMiD~2.0, ODIR-5K) and de-identified by their
providers; no new data were collected. Code, the harmonization rules and the
test-set predictions of the best configuration are available at
\url{https://github.com/asifuddin01/Beyond-Multi-Stage-Pipelines}. The system is a
research prototype, not a medical device.

\IEEEtriggeratref{15}

\end{document}